\documentclass[10pt, a4paper, conference, compsocconf]{IEEEtran}
\ifCLASSINFOpdf
\else
\fi

\usepackage{times}
\usepackage{epsfig}
\usepackage{graphicx}
\usepackage{amsmath}
\usepackage{amssymb}
\usepackage[ruled]{algorithm} 
\usepackage[algo2e]{algorithm2e} 
\usepackage{pbox}
\usepackage{array}
\usepackage{multirow}
\usepackage{color,colortbl}
\usepackage{flushend}

\begin{document}
%
% paper title
% can use linebreaks \\ within to get better formatting as desired
\title{Learning Sparse Adversarial Dictionaries For Multi-Class Audio Classification}

% author names and affiliations
% use a multiple column layout for up to two different
% affiliations

\author{\IEEEauthorblockN{Vaisakh Shaj}
\IEEEauthorblockA{Department of Mathematics\\
Indian Institute Of Space Science And Technology\\
vaisakhs.shaj@gmail.com}
\and
\IEEEauthorblockN{Puranjoy Bhattacharya}
\IEEEauthorblockA{
Intel, Bangalore \\
puranjoy.b@gmail.com}
}

% conference papers do not typically use \thanks and this command
% is locked out in conference mode. If really needed, such as for
% the acknowledgment of grants, issue a \IEEEoverridecommandlockouts
% after \documentclass

% for over three affiliations, or if they all won't fit within the width
% of the page, use this alternative format:
% 
%\author{\IEEEauthorblockN{Michael Shell\IEEEauthorrefmark{1},
%Homer Simpson\IEEEauthorrefmark{2},
%James Kirk\IEEEauthorrefmark{3}, 
%Montgomery Scott\IEEEauthorrefmark{3} and
%Eldon Tyrell\IEEEauthorrefmark{4}}
%\IEEEauthorblockA{\IEEEauthorrefmark{1}School of Electrical and Computer Engineering\\
%Georgia Institute of Technology,
%Atlanta, Georgia 30332--0250\\ Email: see http://www.michaelshell.org/contact.html}
%\IEEEauthorblockA{\IEEEauthorrefmark{2}Twentieth Century Fox, Springfield, USA\\
%Email: homer@thesimpsons.com}
%\IEEEauthorblockA{\IEEEauthorrefmark{3}Starfleet Academy, San Francisco, California 96678-2391\\
%Telephone: (800) 555--1212, Fax: (888) 555--1212}
%\IEEEauthorblockA{\IEEEauthorrefmark{4}Tyrell Inc., 123 Replicant Street, Los Angeles, California 90210--4321}}

% use for special paper notices
%\IEEEspecialpapernotice{(Invited Paper)}

% make the title area
\maketitle

\begin{abstract}
Audio events are quite often overlapping in nature, and more prone to noise than visual signals. There has been increasing evidence for the superior performance of representations learned using sparse dictionaries for applications like audio denoising and speech enhancement. This paper concentrates on modifying the traditional reconstructive dictionary learning algorithms, by incorporating a discriminative term into the objective function inorder to learn class specific adversarial dictionaries that are good at representing samples of their own class at the same time poor at representing samples belonging to any other class. We quantitatively demonstrate the effectiveness of our learned dictionaries as a stand-alone solution for both binary as well as multi-class audio classification problems.

\end{abstract}


% For peer review papers, you can put extra information on the cover
% page as needed:
% \ifCLASSOPTIONpeerreview
% \begin{center} \bfseries EDICS Category: 3-BBND \end{center}
% \fi
%
% For peerreview papers, this IEEEtran command inserts a page break and
% creates the second title. It will be ignored for other modes.
\IEEEpeerreviewmaketitle

\section{Introduction}

The concept of sparsity has received considerable attention in the field of Machine Learning in the past decade. Sparse representations are representations that account for most or all information of a signal with a linear combination of only a few elementary signals, called atoms. The collection of atoms that is used is called a dictionary. Since the bases/atoms being non-orthogonal and not linearly independent of each other as opposed to the traditional orthogonal basis representation, the sparse representation will only recruit those atoms necessary for representing a given input thus resulting in an input-output function whose behavior deviates from being linear and over a wider range of generating element\cite{rubinstein2010dictionaries}. Potentially, this wider range allows more flexibility in signal representation.\par

Representing signals using a set of
learned bases, instead of predefined bases(DCT,wavelet etc), has led to state-of-the-art results mostly in image processing/computer vision tasks such as denoising, inpainting and classification.\par For audio signals, sparse representations have been successfully used for source separation by expressing a signal that is a mixture of multiple sources with a sparse representation, using a dictionary for each
underlying source. However there have been fewer attempts to adapt this model for audio classification, for which we believe sparse representations are a natural fit and have immense possibilities considering the overlapping nature of audio signals. \par 

In this research work, we will be learning class specific adversarial dictionaries that can be used as a stand alone solution for audio classification tasks. We derive
a method to solve the optimization problem, with an adversarial loss in addition to the standard reconstruction loss in the objective function, inorder to learn dictionaries that emphasize inter-class scatter while keeping the intra-class scatter small resulting in enhanced classification performance. We also propose a direct method for training adversarial dictionaries that can be used in the multi-class classification tasks.  

%------------------

%-------------------------------------------------------------------------
\section{Related Works}

There have been several interesting research works in recent times, in the image processing and computer vision fields, indicating the superior performance of dictionaries constructed via supervised
learning for pattern classification tasks, mostly by incorporating a discriminative criterion into the objective function. These algorithms fall under two categories. The first of these \cite{rodriguez2008sparse,zhang2009learning,grosse2012shift,
mairal2008discriminative,boureau2010learning} treats dictionary learning and classifier training as two separate processes, which uses classifiers like SVM\cite{nair2007function} trained on sparse dictionary based features in the final stage.\par 

More sophisticated approaches \cite{yang2010supervised,mairal2009supervised} unify these two processes into a mixed reconstructive and discriminative formulation. They learn simultaneously an over-complete dictionary and multiple linear
classification models for each class. Supervised sparse representation methods like Discriminative KSVD (DKSVD) \cite{zhang2010discriminative}  and its extension Label Consistent K-SVD (LCKSVD) \cite{jiang2013label}, learn atoms of dictionary based on the traditional K-SVD algorithm. It incorporated the classification error into the objective functions in order to enhance the classification ability of coefficients.\par 

In this paper, we will be learning class specific dictionaries that are adversarial in nature. By adversarial we mean each of these class specific dictionaries is learned in such a manner that apart from doing a good job at representing samples of their own class, they also have to do a poor job at representing samples belonging to any other class. This allows us to learn a structured dictionary where atoms have correspondence to the class labels so that the reconstruction error associated with each class can be used for classification and hence can act as a stand-alone classifier as opposed to  \cite{rodriguez2008sparse,zhang2009learning,grosse2012shift,
mairal2008discriminative,boureau2010learning}, which relies on expensive classifiers like RBF Kernal SVM trained on sparse coefficients for making final predictions. We propose a direct method to train these dictionaries in a multi-class setting, unlike \cite{yang2010supervised,mairal2009supervised}, that deal with multiclass problem into multiple independent binary classification tasks and these might not scale well for large number of classes and are often computationally expensive. Our method is unique as compared to previous approaches as they are designed to capture inter-class differences in a much better manner even for multi-class problem instances.

\section{Adversarial Dictionary Learning}

The idea behind this paper is based on the fact that the atoms that have the most energy contribution for the faithful reconstruction
of the signals might not be the atoms that are important for classification purposes \cite{huang2006sparse}. During training , including a discriminative criterion into the dictionary learning objective might help us in learning atoms which are more discriminative and useful for classification purposes. We chose cross-class reconstruction error as the discriminative criterion in virtue of recent researches\cite{he2016optimization} in the area of speech enhancement
aimed at determining what constitute a ``good" dictionary, which suggest that a 
dictionary which represents speech sparsely ideally should be bad at representing noise sparsely and vice versa. Essentially the dictionaries corresponding to the different sources should be adversarial in nature. \par

Let $Y=[y_1,y_2,...,y_N]$ $ \hspace{0.1cm} \epsilon   R^{m \times N}$ be the set of all training signals , $D_c =[d_{1},. . .,d_{k}]$  \hspace{0.1cm} $\epsilon R^{m\times k}$ 
is the dictionary learned with respect to training samples belonging to class c, with each column representing
a basis vector / ``atom" and $S =[\alpha_{1},. . .,\alpha_{N}]$  \hspace{0.1cm} $\epsilon R^{k\times N}$ are the sparse representations of the signal Y over the dictionary $D_c$. Given a
class specific dictionary $D_c$ containing k bases and
a certain number $L<<k$, an L-subspace of $D_c$ is defined
as a span of a subset of L bases from $D_c$ and our aim is to learn dictionaries such that any sample from a class can be reasonably close to an L-subspace of the associated dictionary while a complementary sample is far from any L-subspace of that dictionary. This is achieved by framing the objective function for each class specific sub-dictionary $D_c$ in the following manner : 

\begin{equation}
\label{DFDLObj}
\begin{split}
D_c^{*}=\underset{D_c}{\text{argmin}}\frac{1}{N}   \|Y_c-D_cS_c\|_{F}^2 - \frac{\rho}{\bar{N}}  \|\bar{Y_c}-D_c\bar{S_c}\|_{F}^2 \\  
 subject \quad to \quad \|\alpha_{i}\|_{0} \leq L 
  ,\forall i =\{1,2,...N\}\\
 \quad \quad \quad ,\text{where} \quad  c= \{ 1,2,3,.,.,C \}
\end{split}
\end{equation}

The sparsity constraint on any code $\alpha_i \hspace{0.1cm} \epsilon R^{k}$ is formulated as
$\|\alpha_i\|_{0} \leq L$ with $L<<k$, where $\|\alpha_i\|_{0}$ is the number of non-zero entries in the code. Here C corresponds to the number of classes in the classification task. $\rho$ is the regularization parameter. $Y_c$,$S_c$ corresponds to matrices of all training samples and sparse codes belonging to class c samples and $\bar{Y_c}$,$\bar{S_c}$ corresponds to matrices of all training samples and sparse codes \textbf{not} belonging to class c. N and $\bar{N}$ are the number of training samples  belonging to class c and \textbf{not} belonging to class c respectively. Here $S$ corresponds to the sparse codes corresponding to all training examples and Y being the matrix of all training examples. For any class c, $ Y= [ Y_c,\bar{Y_c} ] $ and  $ S= [ S_c,\bar{S_c} ] $. Throughout this paper we will be referring to the term $ \|Y_c-D_cS_c\|_{F}^2$ as \textbf{reconstruction error} (reconstruction loss) and the term  $\|\bar{Y_c}-D_c\bar{S_c}\|_{F}^2$ as \textbf{cross reconstruction error} (adversarial loss). \par
An iterative method is used to find the optimal solution for
problem (1). Specifically, the process is iterative by fixing $D_c$
while optimizing $S_c$, $\bar{S_c}$ (class specific sparse coding) and vice versa(class specific dictionary update). The two stages are explained in detail below:

\subsection{\textbf{Class Specific Sparse Coding }}

In this stage, for the class specific dictionary $D_c$ corresponding to class c,  we fix $D_c$ and do a sparse coding over all training samples $ Y= [ Y_c,\bar{Y_c} ] $, by solving, 
\begin{equation}
\begin{split}
\label{lassoSparse}
 \underset{S}{\text{argmin}} \frac{1}{N}  \|Y-D_cS\|_{F} \quad subject \quad to \quad \|\alpha_{i}\|_{0} \leq L \\ 
  \forall i =\{1,2,...N\} 
 \end{split} 
\end{equation}

Here $ S= [ S_c,\bar{S_c} ] $ and the sparsity constraint on the code is given by $\|\alpha_i\|_{0} \leq L$. Either L0\cite{rubinstein2008efficient} or L1\cite{efron2004least} norm based methods can used for inducing sparsity. 
\subsection{\textbf{Class Specific Dictionary Update} }

The class specific dictionary update steps for the binary and multi-class problem instances are explained below :
\subsubsection{\textbf{Binary Classification}}
We start with the simple case of binary classification. The class specific dictionary update stage involves fixing the sparse codes S and solving, 

\begin{equation}
\label{lassoSparse}
\underset{D_c}{\text{argmin}}  \frac{1}{N} \|Y_c-D_cS_c\|_{F}^2 - \frac{\rho}{\bar{N}} \|\bar{Y_c}-D_c\bar{S_c}\|_{F}^2  
\end{equation}

Since for any matrix M,  $ \|M\|_{F}^2 = trace(MM^T) $ the objective function in (3) has an alternative representation as follows ,

\begin{equation}
\begin{split}
 &\underset{D_c}{\text{argmin}}\frac{1}{N}   \|Y_c-D_cS_c\|_{F}^2 - \frac{\rho}{\bar{N}}  \|\bar{Y_c}-D_c\bar{S_c}\|_{F}^2  \\
& \quad =   \underset{D_c}{\text{argmin}} \quad \frac{1}{N} trace \big ( (Y_c-D_cS_c)(Y_c-D_cS_c)^{T} \big )  \\
& \quad \quad \quad \quad -\frac{\rho}{\bar{N}}   trace \big ( (\bar{Y_c}-D_c\bar{S_c})(\bar{Y_c}-D_c\bar{S_c})^{T} \big ) \\
  & \quad  =  \underset{D_c}{\text{argmin}}   - 2 trace \big (  AD_c^T \big ) + trace\big ( D_cBD_c^T \big )  \\
  & \quad \quad \quad \quad \quad \quad \quad , where \quad A=\big ( \frac{1}{N} Y_cS_c^T - \frac{\rho}{\bar{N}}  \bar{Y_c}\bar{S_c}^T \big )  \\
  & \quad \quad \quad \quad \quad \quad \quad \quad \quad \quad B=\big (\frac{1}{N} S_cS_c^T -\frac{\rho}{\bar{N}}  \bar{S_c}\bar{S_c}^T \big ) 
\end{split}
\end{equation}
The objective function in equation 4 is very similar to the
objective function in the dictionary update stage problem in
\cite{mairal2010online} except that it is not guaranteed to be convex. If convex it can be effectively solved using a block coordinate descent update\cite{mairal2010online} with a warm start, for each of the column/atom of the dictionary one by one. It is convex
if and only if B is positive semi-definite. For the adversarial
dictionary learning problem, the symmetric matrix B
is not guaranteed to be so, even when all of its
eigenvalues are real. In the worst case, where B is negative
semidefinite, the objective function becomes concave;
if we apply the same dictionary update algorithm as in \cite {mairal2010online},
it will reach its maximum solution instead of the minimum.
To deal with this situation, we need to
convexify the objective function.\par
We took inspiration from \cite{vu2016histopathological}, where the authors used a similar framework for binary classification of Histopathological Images. If we let $\lambda_1(M) \leq \lambda_2(M) \leq \quad . \quad . \quad .  \leq \lambda_{max} (M) $ to be (real) eigenvalues
of a symmetric matrix M, the positive semidefinite constraint of M is equivalent to the non-negativity constraint of $\lambda_1(F)$. Keeping this in mind, replacing B in (4) with $\bar{B} = B - \lambda_{min}(B)I_k$ , will ensure the convexity , since $\bar{B}$ is guaranteed to PSD , where constants $\lambda_{min}(B)$ and $I_k$ are the minimum eigen value of the matrix B and  the identity matrix respectively. It can be easily shown this procedure doesn't change the overall objective (3), given $||d_i||_{2}=1, \forall i= \{1,2,...,k\}$ , where $d_i$ corresponds to the $i^{th}$ column / atom of the dictionary $D_c$ (assuming that each class specific dictionary comprises of k atoms) . \\

\begin{equation}
\begin{split}
  &   \underset{D_c}{\text{argmin}}  - 2 trace \big (  AD_c^T \big ) + trace\big ( D_c(B-\lambda_{min}(B) I_k )D_c^T \big )  \\
  & \quad \quad = \underset{D_c}{\text{argmin}}   - 2 trace \big (  AD_c^T \big ) + trace\big ( D_cBD_c^T \big ) \\
  & \quad \quad \quad \quad \quad \quad \quad \quad + K trace\big ( D_cD_c^T \big )  \\
   & \quad \quad \quad = \underset{D_c}{\text{argmin}}   - 2 trace \big (  AD_c^T \big ) + trace\big ( D_cBD_c^T \big ) + K'\\
   & \quad \quad \quad \quad \quad \quad \quad \quad \quad  , \text{since} \|d_i\|_{2}=1 , \forall i= \{1,2,...,k\} . \\
   &  \quad \quad \quad \quad \quad \quad \quad \quad \quad  \text{Here K and $K'$ are constants.}    
\end{split}
\end{equation}

\subsubsection{\textbf{Multi Class Classification}}

The natural extension of adversarial dictionary learning algorithm for multi class problems, involves including all the off class examples in the cross reconstruction term as a single batch during each dictionary update step. Here by off class we mean all those training samples that don't belong to the particular class specific dictionary which is being updated. The procedure is explained in Algorithm 1.\\
However updating the class specific dictionary in a manner such that each of these updates will have a mini batch of examples, belonging to a particular off class instead of all the off classes, in the cross reconstruction term was the optimization procedure that we used. The procedure is explained in Algorithm 2.\\
Empirical results in section 4 prove that, Algorithm 2 is a much more effective dictionary update strategy than Algorithm 1, and should be used for multi-class problem instances.\\

\begin{algorithm}[ht]
\SetAlgoLined
  \caption{Adversarial DL Algorithm 1 - Multi Class Problems}
1.Initialize the dictionary $D = [D_1, . . .. , D_C]$  ,where Di = [$d^i_1$,. . .,$d^i_k$] $\epsilon \hspace{.1cm} R^{m \times k}$ is the class specific dictionary and $Y_i \hspace{.1cm}$ $\epsilon R^{m \times N_c}$ is the matrix of all training samples belonging to any class $i \hspace{.1cm} \epsilon \{1,2,3,....C\}$. 

2.\For{Each class i=1 to C }{

\While{not converged}{
    
  \textbf{a.} Fix $D_i$ and update $S=[S_1,S_2,....,S_C]$ by solving \hspace{.1cm} (2);\\
		     \textbf{b.} $A=\frac{1}{N_i} Y_{i}S_{i}^{T} - \frac{\rho}{N_j} \sum_{j} Y_{j}S_{j}^{T}$ \\
  			    $B =\frac{1}{N_i} S_{i}S_{i}^{T} - \frac{\rho}{N_j} \sum_{j} S_{j}S_{j}^{T}$, \\ 
  			    \hspace{1cm}$ \forall j    \big ( j \epsilon \{ 1,2, . . ., C \}\quad \cap \quad j \neq i \big )  $  \\
  			    \\
   			 \textbf{c.} Update each column of $D_{i}$ using block coordinate descent\cite{mairal2010online}, so that updated 
   			 $ D_i=\underset{D_i}{\text{argmin}} \quad  - 2 trace \big (  AD_i^T \big ) + trace\big ( D_i\bar{B}D_i^T \big ) $\\ , subject to: $\|d^i_x\|_{2}=1$ , $x = \{1,2,3,....k\}$  , where $\bar{B} = B - \lambda_{min}(B)I_k$ 
            
	}
		}

\end{algorithm}

\begin{algorithm}[ht]
\SetAlgoLined
  \caption{Adversarial DL Algorithm 2 - Multi Class Problems}
1.Initialize the dictionary $D = [D_1, . . .. , D_C]$  ,where Di = [$d^i_1$,. . .,$d^i_k$] $\epsilon \hspace{.1cm} R^{m \times k}$ is the class specific dictionary and $Y_i \hspace{.1cm}$ $\epsilon R^{m \times N_c}$ is the matrix of all training samples belonging to any class $i \hspace{.1cm} \epsilon \{1,2,3,....C\}$ .

2.\For{Each class i=1 to C }{
\While{not converged}{

		\textbf{a.} Fix $D_i$ and update $S=[S_1,S_2,....,S_C]$ by solving \hspace{.1cm} (2);

		\For{Each class j $ \epsilon \{ 1,2, . .. ., C \} $ , except i}{

  			 \textbf{a.} $A=\frac{1}{N_i} Y_{i}S_{i}^{T} -\frac{\rho}{N_j} Y_{j}S_{j}^{T}$ \\
  			    $\quad B=\frac{1}{N_i} S_{i}S_{i}^{T} - \frac{\rho}{N_j} S_{j}S_{j}^{T}$ \\
  			    \\
   			 \textbf{b.} Update each column of $D_{i}$ using block coordinate descent\cite{mairal2010online}, so that updated
   			 $D_i =\underset{D_i}{\text{argmin}}-2 trace \big (  AD_i^T \big )+trace\big ( D_i\bar{B}D_i^T \big ) $ \\
   			 , subject to: $\|d^i_x\|_{2}=1$ , $x = \{1,2,3,....k\}$  , where $\bar{B} = B - \lambda_{min}(B)I_k$ 
            
		}
		}
		}
\end{algorithm}

\section{Experimental validation}

We describe in this section the results obtained using our sparse dictionary based classification framework, with (dubbed Adversarial DL) and without the discriminative criterion (dubbed Reconstructive DL) in the objective function, for various Binary and Multi-Class Audio Classification Problems. The experiments are conducted on two datasets , IEEE DCASE 2013 dataset and GTZAN Music Genre Classification Dataset. For both datasets the value of regularization parameter $\rho$ is selected based on a grid-like search over
$\rho$ (between 0.00001 and 0.1, on a log scale). \\
\subsection{\textbf{IEEE 2013 DCASE Dataset}}

The dataset consists of 3 subsets (for development, training, and testing), of which the training set will contain instantiations of individual events for every class. The developement and testing datasets, consists of 1 minute recordings each of every-day audio events in a number of office environments. The audio events for these recordings was annotated to eliminate silent frames and events corresponding to each of the classes were isolated. We chose 2 classes , ClearThroat and Cough, from the dataset to check the performance of our system as a binary classifier. These were chosen since the classes were highly correlated and sounds very identical even to human ears.  4 classes were chosen to validate the performance of our classification framework in the multi class setting, the classes being ClearThroat, Cough, Doorslam and Drawer.\par
The pre-processing involves generating a normalized magnitude spectrogram output from the wav files such that output levels range from 0 to about 120 for TF bins, roughly corresponding to the 120 dB audio dynamic range. We first divide an utterance into a number of overlapping, fixed-length windows, with the window length equal to the dictionary atom size. In this work, we have kept the window width to 50 frames using a window shift of $\Delta$ frames. Larger values of $\Delta$ reduce computational effort but might decrease representational accuracy\cite{gemmeke2011exemplar}. For this dataset, we keep the window shift constant at $\Delta = 1$ frame. The sparse coefficients are computed using OMP\cite{rubinstein2008efficient} with a fixed number of coefficients, where the maximal number of coefficients was set to two in our experiments both during the training as well as testing phase(as this was giving good performance with less computational cost).   \\

\begin{table}[!t]
\caption{Test Accuracy($\%$) Binary(Task 1) and 4-way classification(Task 2) on IEEE DCASE Event Dataset}
\begin{tabular}{|c|c|c|l|} \hline
&Task 1  &Task 2\\ \hline
\texttt{Reconstructive DL  } &68.4 & 77.91\\
\texttt{Adversarial DL-1 ($\rho = .001$)   } &\textbf{74.81} & 80.71\\
\texttt{Adversarial DL-2 ($\rho = .001$)  } &\textbf{74.81} & \textbf{83.02}\\
\texttt{D-KSVD  } &56.58 & {56.71} \\

\texttt{D-KSVD + RBF SVM  } &69.56 & {61.12}\\
\texttt{LC-KSVD  } &60.86 & {55.31} \\

\texttt{LC-KSVD + RBF SVM  } &73.90 & {64.67} \\
\hline\end{tabular}
\end{table}

\begin{figure}[ht]
\begin{center}
\includegraphics[width=0.5\textwidth]{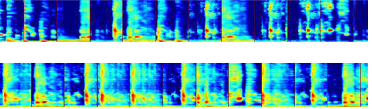}
\caption{Examples of learned bases with discriminative criterian (top) and without discriminative criterian(bottom) for the cough class from DCASE 2013 dataset .}
\includegraphics[width=0.5\textwidth]{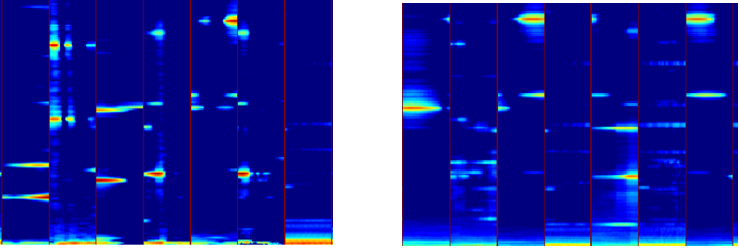}
\caption{Examples of learned bases with discriminative criterian (left) and without discriminative criterian(right) for the alert class from DCASE 2013 dataset .}
\label{fig:spiral}
\end{center}
\end{figure}

Once the representations were learned, any test signal y is sparse coded over each of the class specific dictionaries $D_c$ to obtain $\alpha_c$, where $c \hspace{.1cm} \epsilon \{1,2,....,C\}$ and reconstruction error is used as an index for classfication.

$$ identity(y) =  \underset{c}{\text{argmin}} \mathbf{ \|y-D_c \alpha_c\|_{2}^2} $$

Table 1 shows the results we obtained for the binary and multi-class classification tasks using various supervised dictionary learning approaches. The  D-KSVD and LC-KSVD use the linear predictive classifier to predict the label. We also hybridized the D-KSVD and LC-KSVD by treating dictionary learning and classifier training as two separate processes using an RBF Kernal SVM. From the table it is clear that the dictionaries learned with the discriminative criterion of maximizing the cross reconstruction error gave superior classification performance even without the use of any additional classifier. Also, Algorithm 2, for multi-class problem instances were learning much better dictionaries than Algorithm 1. Figures 1 and 2 compare the representations learned using Reconstructive DL and Adversarial DL for selected classes. It is clear from the figure that the later produced bases with higher details than the former which in turn produced bases that are blurred versions of the actual training patches. 
\begin{figure}[ht]
\begin{center}
\includegraphics[width=0.5\textwidth]{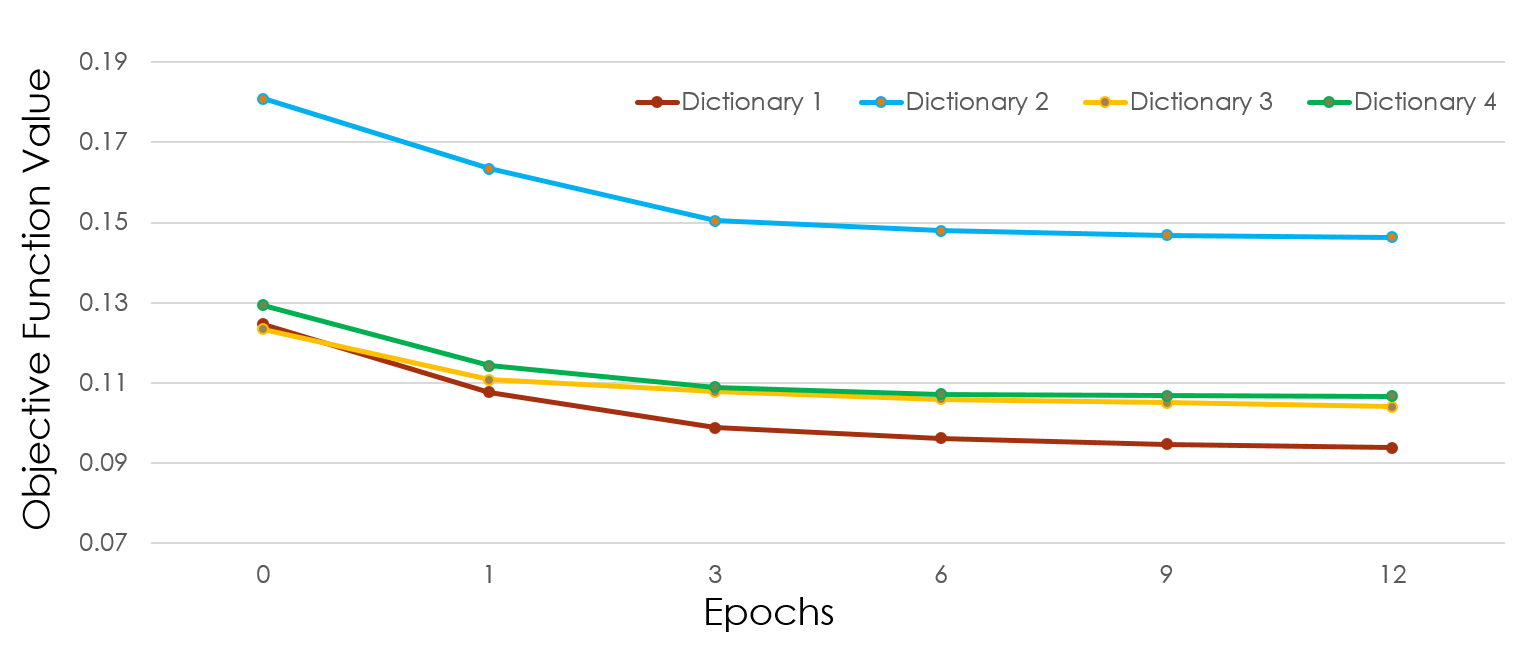}
\caption{The convergence of Adversarial DL-2 on IEEE DCASE Dataset in Multi-Class Setting for each of the 4 Class Specific Dictionaries.}\label{fig:spiral}
\end{center}
\end{figure}
\subsection{\textbf{Music Genre Classification - GTZAN Dataset}}

The second dataset, abbreviated as GTZAN consists of 10 genre classes. Each genre class contains 100 audio recordings 30 sec long.

A stratified 10-fold cross-validation is employed for experiments conducted on the GTZAN dataset and the average accuracy is reported. Thus each training set consists of 900 audio files and the test set containing 100 audio files.

The pre-processing stage is almost similar to the previous sections except that dictionaries were learned in the log magnitude spectrograms, with windows of length 50 frames. Once the representations were learned, the test signals are sparse coded over each of the class specific dictionaries corresponding to the particular classes and reconstruction error is used as an index for classfication. Inorder to find the class of a particular 30 second length test signal, majority voting was employed. The results with non aggregated features are shown in Table 2.

\begin{table}[!t]
\caption{Test Accuracy($\%$) for 10-way music genre classification on GTZAN Dataset}
\begin{tabular}{|c|c|} \hline
Model &Accuracy\\ \hline
\texttt{ Reconstructive DL  } &$74 $ \\
\texttt{ Adversarial DL-1 \small($\rho = 0.0001$ )  } & $75.5 $\\
\texttt{ Adversarial DL-2 \small($\rho = 0.001$ )  } &\textbf{$\textbf{78} $} \\ \hline
\texttt{ MFCC + SVM   } & 63 \\
\texttt{GMM+MFCC(m5,v5) + other }  &61 \\
\texttt{LDA+MFCC(m5,v5) + other } &71 \\
\texttt{AdaBoost + many features }   &\textbf{83}  \\ \hline
\texttt{ DBN Layer 1 features + RBF SVM   } &73.5 \\
\texttt{ DBN Layer 2 features + RBF SVM  } &\textbf{77} \\ 
\texttt{ DBN Layer 3 features + RBF SVM  } &73.5 \\
\texttt{ CNN  } &72 \\ \hline\end{tabular}
\end{table}

\begin{figure}[ht]
\begin{center}
\includegraphics[width=0.5\textwidth]{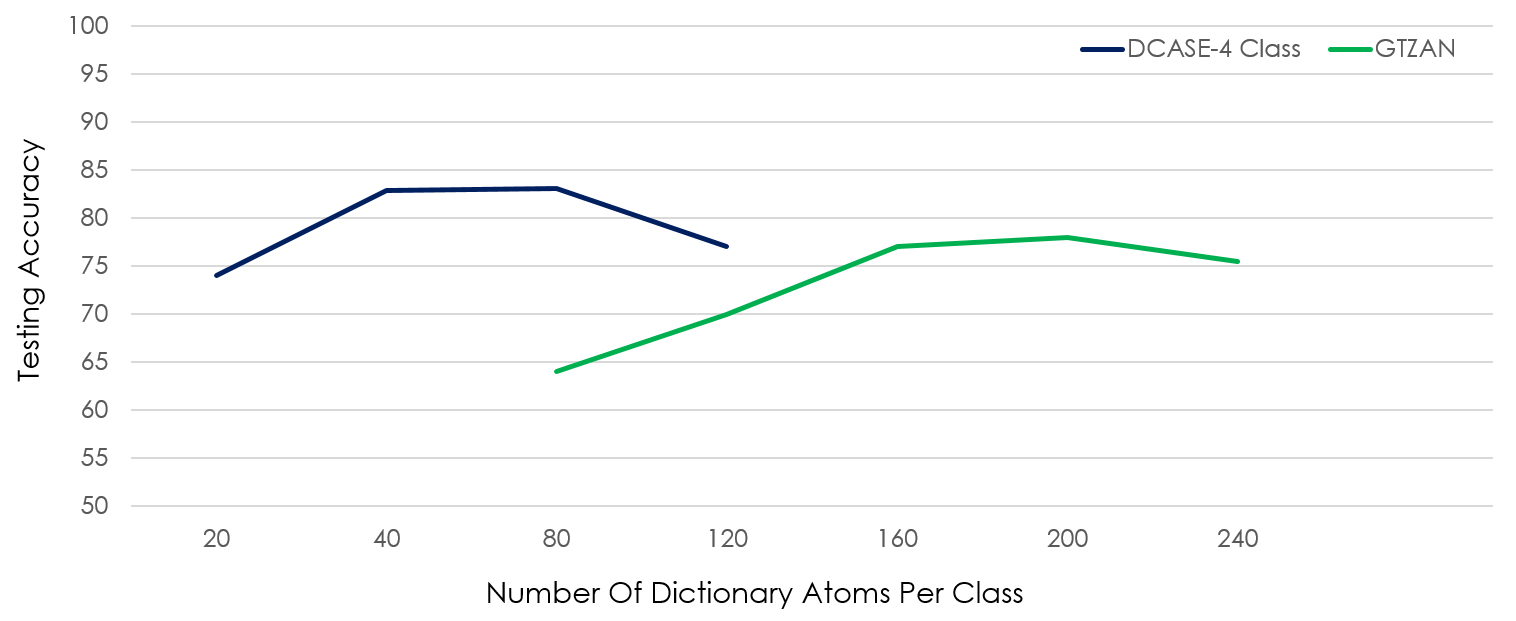}
\caption{Recognition results with different
dictionary sizes. From the figure we can infer that for discriminative tasks, increasing the number of atoms beyond a threshold is
likely to lead to overfitting, and smaller values are preferred unlike reconstructive tasks where increasing the number of atoms leads to better capabilities. }\label{fig:spiral}
\end{center}
\end{figure}

The accuracy of $78 \%$ obtained by our method exceeds the accuracy
of $76 \%$, pointed out by \cite{mckay2006musical}, obtained when
human beings correctly classify music songs. It's clear from Table 2 that our approach performed better than most systems that pair signal processing feature extractors  \cite{li2003factors,tzanetakis2002musical} with standard classifiers such as SVMs, Nearest Neighbors or Gaussian Mixture Models. Considering the fact that we learned dictionaries on generic log magnitude spectrograms and have not resorted to any music specific pre-processing as in \cite{henaff2011unsupervised}, which performed unsupervised feature learning in a transductive setting on the entire dataset
(which includes the training and test sets during feature learning), our system report one of the highest accuracies using sparse dictionary based techniques for GTZAN dataset.  We also compared the results with some of the state of the art deep learning techniques; Deep Belief Networks(DBN) with non-aggregated feature\cite{hamel2010learning} and a Convolutional Neural Network based approach(CNN)\cite{nakashika2012local}. Its clear from Table 2 that, the adversarial dictionary based classifier is a powerful tool for audio classification tasks and can be used as a standalone classifier. \par
Here we have not used aggregation of features as in \cite{bergstra2006aggregate}, which is considered as a good method for music information retrieval tasks since it was not giving further improvements in our preliminary experiments. Techniques to incorporate this in our framework requires further investigation as this may enhance the capability to capture temporal dependencies between frames.

\section{Conclusion And Future Work}
We argued that sparse dictionaries can and should be used as
stand-alone non-linear classifiers alongside other standard
and more popular classifiers, instead of merely being considered as simple feature extractors. We evaluated a discriminative training objective that is more appropriate to train sparse dictionaries for classification problems both in binary as well as mutli-class settings. These adversarial versions of sparse dictionaries integrate the process of discovering features of inputs with their use
in classification, without relying on a separate classifier. Extending this framework for overlapping audio events(multi-label problems) and to other type of data including images in the multi-class setting will be taken up as a future work.

%------------------------------------------------------------------------

{\small
\bibliographystyle{ieee}
\bibliography{egbib}
}

\end{document}